\pdfoutput=1

\documentclass[11pt,x11names]{article}

\usepackage[]{EMNLP2023}

\usepackage{times}
\usepackage{latexsym}

\usepackage[T1]{fontenc}

\usepackage[utf8]{inputenc}

\usepackage{microtype}
\usepackage{microtype}
\usepackage{times}
\usepackage{latexsym}
\usepackage{amssymb}
\usepackage{amsfonts}
\usepackage{amsmath} 
\usepackage{amsthm} 
\usepackage{booktabs}
\usepackage{enumerate}
\usepackage{enumitem}
\usepackage{graphicx}
\usepackage{subfigure}
\usepackage{xspace}
\usepackage{float}
\usepackage{bbm}
\usepackage{bm}
\usepackage{multirow}
\usepackage{booktabs}
\usepackage{color}
\usepackage{framed}
\usepackage{stfloats}
\usepackage{iitem}
\usepackage{makecell}
\usepackage{float}
\usepackage{placeins}
\usepackage{afterpage}
\usepackage{bbding}
\usepackage[normalem]{ulem}
\usepackage{soul}
\usepackage{xcolor}
\usepackage{color}
\usepackage{colortbl}
\usepackage{algorithm}
\usepackage{algorithmic}
\usepackage{array}
\usepackage{hyperref}
\usepackage{listings}
\usepackage{lipsum,mwe,cuted}

\usepackage{url}
\newcommand{\paratitle}[1]{\vspace{1.5ex}\noindent\textbf{#1}}
\newcommand{\ie}{\emph{i.e.,}\xspace}

\newcommand{\eg}{\emph{e.g.,}\xspace}

\newcommand{\ignore}[1]{}
\newcommand{\tabincell}[2]{\begin{tabular}{@{}#1@{}}#2\end{tabular}}

\usepackage{xcolor}
\definecolor{gold}{RGB}{205,133,63}
\definecolor{fGreen}{RGB}{34,139,34}
\definecolor{tOrange}{RGB}{255,165,0}
\definecolor{tBlue}{RGB}{135,206,250}
\definecolor{tPink}{RGB}{255,204,204}
\definecolor{tGreen}{RGB}{205,230,199}
\definecolor{tGold}{RGB}{255,215,0}
\newcolumntype{P}[1]{>{\centering\arraybackslash}p{#1}}
\newcolumntype{M}[1]{>{\centering\arraybackslash}m{#1}}

\usepackage{inconsolata}

\title{HaluEval: A Large-Scale Hallucination Evaluation Benchmark\\ for Large Language Models}

\author{
	Junyi Li\textsuperscript{\rm{1,3,4}}\thanks{\ \ Equal contributions}, 
	Xiaoxue Cheng\textsuperscript{\rm{1}}\footnotemark[1],
        Wayne Xin Zhao\textsuperscript{\rm{1,4}\thanks{\ \ Corresponding author}\ }, 
	Jian-Yun Nie\textsuperscript{\rm{3}} {\rm and} 
	Ji-Rong Wen\textsuperscript{\rm{1,2,4}} \\
	\textsuperscript{1}Gaoling School of Artificial Intelligence, Renmin University of China \\
	\textsuperscript{2}School of Information, Renmin University of China \\
	\textsuperscript{3}DIRO, Universit\'{e} de Montr\'{e}al \\
	\textsuperscript{4}Beijing Key Laboratory of Big Data Management and Analysis Methods \\
	\texttt{lijunyi@ruc.edu.cn} \quad
        \texttt{chengxiaoxue3@gmail.com} \quad
	\texttt{batmanfly@gmail.com} \\
}

\begin{document}
\maketitle
\begin{abstract}
Large language models (LLMs), such as ChatGPT, are prone to generate hallucinations, \ie content that conflicts with the source or cannot be verified by the factual knowledge. To understand \emph{what types of content} and \emph{to which extent} LLMs  are apt to hallucinate, we introduce the \textbf{\underline{Hal}}l\textbf{\underline{u}}cination \textbf{\underline{Eval}}uation benchmark for Large Language Models (\textbf{HaluEval}), a large collection of 
generated and human-annotated hallucinated samples for evaluating the performance of LLMs in recognizing hallucination. To generate these samples automatically, we propose a two-stage framework, \ie \emph{sampling-then-filtering}. 
Besides, we hire some human labelers to annotate the hallucinations in ChatGPT responses. The empirical results suggest that ChatGPT is likely to generate hallucinated content related to specific topics by fabricating unverifiable information (\ie about $19.5\%$ responses).
Moreover, existing LLMs face great challenges in recognizing the hallucinations in texts.  However, our experiments also prove that providing external knowledge or adding reasoning steps can help LLMs recognize hallucinations. Our benchmark can be accessed at \url{https://github.com/RUCAIBox/HaluEval}.
\end{abstract}

 \section{Introduction}
\label{sec-intro}

The advent of large language models (LLMs)~\cite{LLM-survey} has ushered in a paradigm shift in natural language processing (NLP), making unprecedented progress in  text generation and understanding~\cite{gpt3,PLM4TG_survey}. The remarkable language ability makes LLMs core in a number of products with millions of users, such as the coding assistant Copilot and recent ChatGPT.

Despite these prominent capabilities of LLMs trained on large text corpus, recent work has shown that LLMs are prone to suffer from \emph{hallucination generations} across various applications~\cite{hallucination-survey,Bang-arxiv-2023-A,Sun2023ContrastiveLR}, where the generated content is either in conflict with existing source or cannot be verified by the available  knowledge resources. The issue of hallucination makes the deployment of LLMs potentially risky in real-world applications. Most exiting work mainly focuses on investigating the causes of hallucination for specific tasks and small language models~\cite{summarization-hallucination,QA-Hallucination,dialogue-hallucination}. However, it still remains unclear \emph{what types of content} and \emph{to which extent} LLMs tend to hallucinate. 

\begin{table}[t]
\renewcommand\arraystretch{1.1}
\setlength\tabcolsep{3pt}
	\centering
	\small
	\begin{tabular}{r p{0.36\textwidth}}
		\toprule[1pt]
		\textbf{User Query} & {Retrieve the oldest photo of a cat} \\
            \midrule[0.5pt]
            \textbf{ChatGPT} & There is no definitive answer to this question\\ 
            \multicolumn{2}{p{0.465\textwidth}}{as ``the oldest photo of a cat'' is subjective and varies based on different sources and definitions. However, \textcolor{fGreen}{\textbf{one candidate for the oldest cat photo is the daguerreotype ``Cat with Blue Ribbon,'' taken by photographer Antoine Claudet in 1840.}} The photo depicts a cat sitting on a table, gazing directly at the camera, and wearing a blue ribbon tied around its neck. [...]} \\
		\bottomrule[1pt]
	\end{tabular}%
	\caption{An example from Alpaca~\cite{alpaca} showing that ChatGPT might generate hallucinated contents (\textcolor{fGreen}{\textbf{green}}) that cannot be verified by existing source.}
        \label{tab-example}%
\end{table}

\begin{figure*}[tb]
	\centering
	\includegraphics[width=0.95\textwidth]{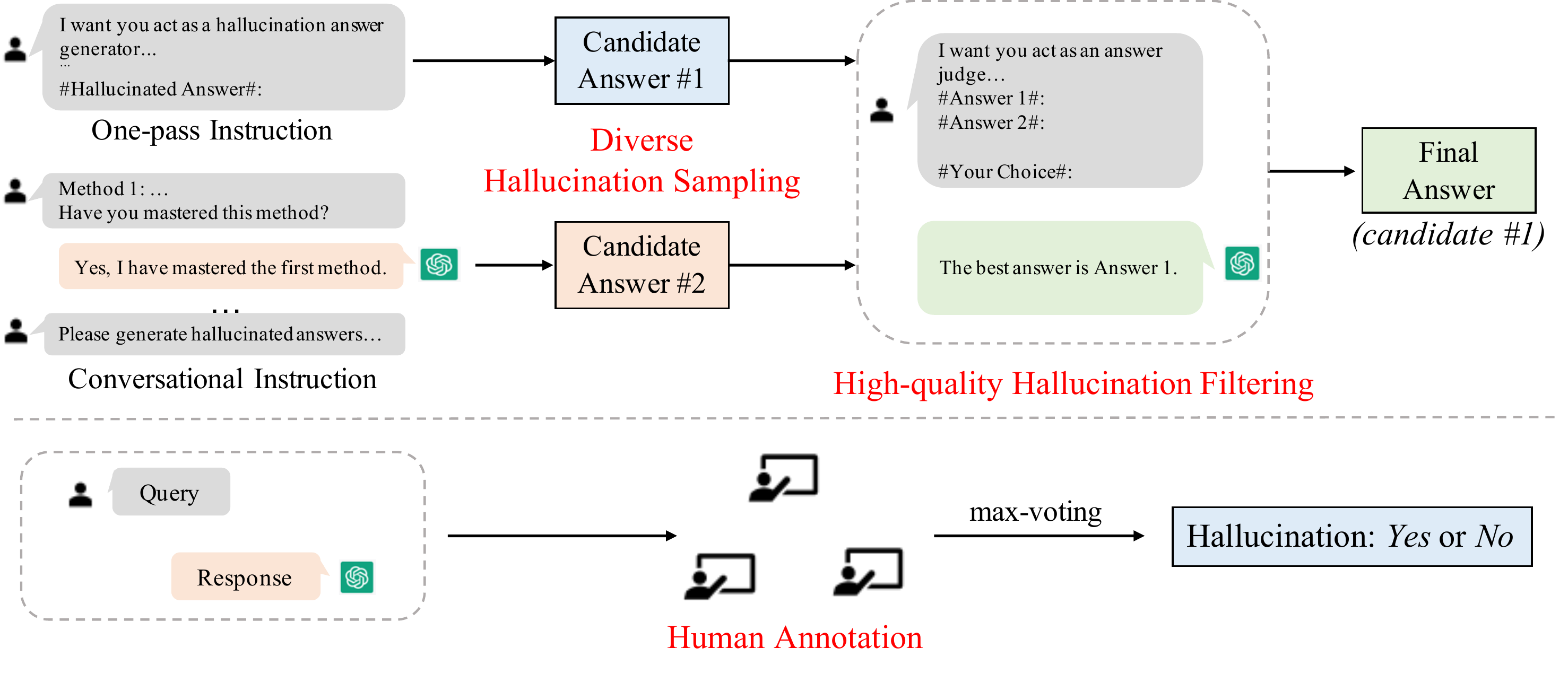}
	\caption{Construction pipeline of HaluEval, including automatic generation (top) and human annotation (bottom).}
	\label{fig-model}
\end{figure*}

To facilitate research in this direction, we present the \textbf{\underline{Hal}}l\textbf{\underline{u}}cination \textbf{\underline{Eval}}uation benchmark for Large Language Models (\textbf{HaluEval}): a large collection of 35,000 hallucinated/normal samples for LLMs analysis and evaluation. HaluEval includes 5,000 \emph{general} user queries with ChatGPT responses and 30,000 \emph{task-specific} examples from three tasks, \ie question answering, knowledge-grounded dialogue, and text summarization. The construction pipeline of HaluEval is depicted in Figure~\ref{fig-model}. For general user queries, we adopt the $52$K instruction tuning dataset from Alpaca~\cite{alpaca} for human annotation. To further screen out user queries where LLMs are most likely to produce hallucinations, we use ChatGPT to sample three responses for each query and only retain $5,000$ queries with the lowest similarity among three responses. According to recent work~\cite{selfcheckgpt}, hallucinations are likely to appear in diverged and conflicting responses of LLMs. Based on the filtered user queries and ChatGPT responses, we invite human labelers to annotate whether the response contains hallucinated information and mark corresponding spans. As shown in Table~\ref{tab-example}, for the user query ``\emph{Retrieve the oldest photo of a cat}'', the response generated by ChatGPT contains unverifiable information. These human-annotated queries and responses can be used to analyze what types of content LLMs tend to hallucinate and further conceive effective methods to alleviate it.

Furthermore, for the task-specific examples, we design an automatic two-stage approach to generate hallucinated samples. First, based on existing task datasets (\eg HotpotQA) as seed data, we employ ChatGPT to generate hallucinated samples with two styles of task-specific instructions, \ie one-pass and conversational. We expect that these two methods will generate diverse hallucinated samples from different aspects. Second, to select the most plausible and difficult hallucinated sample for LLMs evaluation, we elaborate the filtering instruction enhanced by ground-truth examples and leverage ChatGPT for sample selection. Through the proposed sampling-then-filtering approach, we can generate a hallucinated counterpart for each specific task example. These hallucinated samples are designed to challenge the ability of LLMs in hallucination recognition and analyze the information blind spots of LLMs.

To better understand the performance of LLMs in HaluEval, we conduct experiments with several existing powerful LLMs (\eg ChatGPT, GPT-3). Our key findings can be summarized as follows:

\textbullet~First, ChatGPT is likely to generate hallucinated content by fabricating unverifiable information in its responses (\ie about $19.5\%$ responses). The hallucinated texts from ChatGPT cover topics including language, climate, and technology.

\textbullet~Second, existing LLMs face significant challenges to identify the hallucinations in the generated text, even for ChatGPT which is used to generate these hallucinated samples (\eg only $62.59\%$ accuracy for ChatGPT in question answering).

\textbullet~Finally, the deficient performance of LLMs in recognizing hallucinations can be improved by providing explicit knowledge and adding intermediate reasoning steps. While, contrasting hallucinated samples with ground-truth makes LLMs more confused and leads to worse performance.







\section{The HaluEval Benchmark}

As the goal of HaluEval is to understand \emph{what types of content} and \emph{to which extent} LLMs tend to hallucinate, the benchmark contains a myriad of correct samples and their hallucinated counterparts. This collection is created via two ways, \ie automatic generation and human annotation.

\subsection{Automatic Generation}
\label{sec-generation}

Our generation pipeline includes two steps: 1) 
diverse hallucination sampling, and 2) high-quality hallucination filtering. We employ ChatGPT to execute the creation pipeline automatically.

\paratitle{Diverse Hallucination Sampling.} Since a factual text can be hallucinated from different aspects, we propose two different hallucination sampling methods to generate diverse samples. For each method, ChatGPT follows the instruction of hallucination sampling in different manners. As shown in Figure~\ref{fig-model}, the first method adopts a \emph{one-pass} instruction following schema, where we directly feed the complete instruction (Table~\ref{tab:qa-gen-instruction}) into ChatGPT and generate a hallucinated answer. On the other hand, the second method uses a \emph{conversational} schema, where we teach ChatGPT to successively learn part of the instruction and make sure it has mastered. Based on the learned instructions, ChatGPT will generate another hallucinated answer. Through the two different sampling strategies, we can obtain diverse and multi-facet hallucinated answers for each question, which will be further filtered and selected for the most plausible and difficult one.


\begin{table*}[t]
	\small
	\centering
	\begin{tabular}{p{0.97\textwidth}}
		\toprule
            \rowcolor{LightSkyBlue1}{\fontfamily{ppl}\selectfont I want you act as a hallucination answer generator. Given a question, right answer, and related knowledge, your objective is to write a hallucinated answer that sounds plausible but is factually incorrect. You SHOULD write the hallucinated answer using the following method (each with some examples):} \\
            \specialrule{0em}{1pt}{1pt}
            \rowcolor{tPink}{\fontfamily{ppl}\selectfont You are trying to answer a question but there is a factual contradiction between the answer and the knowledge. You can fabricate some information that does not exist in the provided knowledge.} \\
            \specialrule{0em}{1pt}{1pt}
            \rowcolor{tGreen}{\fontfamily{ppl}\selectfont \textbf{\#Knowledge\#:} The nine mile byway starts south of Morehead, Kentucky and can be accessed by U.S. Highway 60. Morehead is a home rule-class city located along US 60 (the historic Midland Trail) and Interstate 64 in Rowan County, Kentucky, in the United States.} \\
            \rowcolor{tGreen}{\fontfamily{ppl}\selectfont \textbf{\#Question\#:} What U.S Highway gives access to Zilpo Road, and is also known as Midland Trail?} \\
            \rowcolor{tGreen}{\fontfamily{ppl}\selectfont \textbf{\#Right Answer\#:} U.S. Highway 60 } \\
            \rowcolor{tGreen}{\fontfamily{ppl}\selectfont \textbf{\#Hallucinated Answer\#:} U.S. Highway 70} \\
            \specialrule{0em}{1pt}{1pt}
            \rowcolor{tPink}{\fontfamily{ppl}\selectfont You are trying to answer a question but you misunderstand the question context and intention.} \\
            \specialrule{0em}{1pt}{1pt}
            \rowcolor{tGreen}{\fontfamily{ppl}\selectfont <Demonstrations>} \\
            \specialrule{0em}{1pt}{1pt}
            \rowcolor{tPink}{\fontfamily{ppl}\selectfont You are trying to answer a question but the answer is too general or too specific to answer the question at an appropriate level of specificity.} \\
            \specialrule{0em}{1pt}{1pt}
            \rowcolor{tGreen}{\fontfamily{ppl}\selectfont <Demonstrations>} \\
            \specialrule{0em}{1pt}{1pt}
            \rowcolor{tPink}{\fontfamily{ppl}\selectfont You are trying to answer a question but the answer cannot be inferred from the knowledge. You can incorrectly reason with the knowledge to arrive at a hallucinated answer.} \\
            \specialrule{0em}{1pt}{1pt}
            \rowcolor{tGreen}{\fontfamily{ppl}\selectfont <Demonstrations>} \\
            \specialrule{0em}{1pt}{1pt}
            {\fontfamily{ppl}\selectfont You should try your best to make the answer become hallucinated. \#Hallucinated Answer\# can only have about 5 more words than \#Right Answer\#. } \\
            \\
            {\fontfamily{ppl}\selectfont \textbf{\#Knowledge\#:} <insert the related knowledge>} \\
            {\fontfamily{ppl}\selectfont \textbf{\#Question\#:} <insert the question>} \\
            {\fontfamily{ppl}\selectfont \textbf{\#Right Answer\#:} <insert the right answer to the question>} \\
            {\fontfamily{ppl}\selectfont \textbf{\#Hallucinated Answer\#:}} \\
            \bottomrule
	\end{tabular}
	\caption{Instruction of hallucination sampling for question answering. The \colorbox{LightSkyBlue1}{blue} text denotes the intention description, the \colorbox{tPink}{red} text denotes the hallucination pattern, and the \colorbox{tGreen}{green} text denotes the hallucination demonstration.} 
	\label{tab:qa-gen-instruction}
\end{table*}

\paratitle{Instruction Design.} In our approach, the key is to design an effective instruction for ChatGPT to generate hallucinated samples. In our design, the  hallucination sampling instruction consists of three important parts, including \emph{intention description}, \emph{hallucination pattern}, and \emph{hallucination demonstration}, which have been shown in Table~\ref{tab:qa-gen-instruction}. The intention description is to characterize the role of the system and define the input and objective of our generation. To control the type and quality of hallucinated samples, we introduce the hallucination pattern and demonstration, which are related to the seed task (\eg QA in Table~\ref{tab:qa-gen-instruction}). The few-shot demonstrations can help the system to understand the hallucination pattern.
In this paper, we automatically generate hallucinated samples for three tasks, \ie question answering, knowledge-grounded dialogue, and text summarization. Specifically, we consider four types of hallucination patterns for question answering (\ie \emph{comprehension}, \emph{factualness}, \emph{specificity}, and \emph{inference})~\cite{QA-Hallucination}, three types of hallucination patterns for knowledge-grounded dialogue (\ie \emph{extrinsic-soft}, \emph{extrinsic-hard}, and \emph{extrinsic-grouped})~\cite{dialogue-hallucination}, and three types of hallucination patterns for text summarization (\ie \emph{factual}, \emph{non-factual}, and \emph{intrinsic})~\cite{summarization-hallucination}. For these three tasks, we first randomly sample $30,000$ instances from the training set of HotpotQA~\cite{yang2018hotpotqa}, OpenDialKG~\cite{Moon2019opendialkg}, and CNN/Daily Mail~\cite{SeeLM17cnndm}, and then generate their hallucinated examples. The hallucination sampling instructions for dialogue and summarization can be found in Table~\ref{tab:dialogue-gen-instruction}-\ref{tab:summary-gen-instruction} in the Appendix ~\ref{app-sampling}.

\paratitle{High-quality Hallucination Filtering.} To construct a challenging benchmark for LLMs, we aim to select the most plausible and difficult hallucinated samples from the above two sampling methods. As shown in Table~\ref{tab:qa-filter-instruction}, we design the instruction of hallucination filtering enhanced by ground-truth answers to select the best answer from two hallucinated candidates. In the instruction of filtering, the demonstration includes the ground-truth correct answer (\eg U.S. Highway 60) and a hallucinated counterpart (\eg U.S. Highway 70). While, in the test example, we input two hallucinated answers. Following the demonstrations, we expect ChatGPT to select one of the hallucinated answers that is the most plausible and closest to the right answer. Finally, the selected hallucinated sample is hard to be identified, which are further used to evaluate LLMs in hallucination recognition. The instructions of hallucination filtering for dialogue and summarization are shown in Table~\ref{tab:dialogue-filter-instruction}-\ref{tab:summary-filter-instruction} in the Appendix~\ref{app-filtering}.

Through the \emph{sampling-then-filtering} process, we end up generating a total of $30,000$ hallucinated samples for the three tasks. Our approach can also be adapted to other tasks and datasets.


\begin{table*}[t]
	\small
	\centering
	\begin{tabular}{p{0.97\textwidth}}
		\toprule
            \rowcolor{LightSkyBlue1}{\fontfamily{ppl}\selectfont I want you act as an answer judge. Given a question, two answers, and related knowledge, your objective is to select the best and correct answer without hallucination and non-factual information. Here are some examples:} \\
            \specialrule{0em}{1pt}{1pt}
            \rowcolor{tGreen}{\fontfamily{ppl}\selectfont \textbf{\#Knowledge\#:}The nine mile byway starts south of Morehead, Kentucky and can be accessed by U.S. Highway 60. Morehead is a home rule-class city located along US 60 (the historic Midland Trail) and Interstate 64 in Rowan County, Kentucky, in the United States.} \\
            \rowcolor{tGreen}{\fontfamily{ppl}\selectfont \textbf{\#Question\#:} What U.S Highway gives access to Zilpo Road, and is also known as Midland Trail?} \\
            \rowcolor{tGreen}{\fontfamily{ppl}\selectfont \textbf{\#Answer 1\#:} U.S. Highway 60 \textcolor{red}{(right answer)}} \\
            \rowcolor{tGreen}{\fontfamily{ppl}\selectfont \textbf{\#Answer 2\#:} U.S. Highway 70 \textcolor{red}{(hallucinated answer)}} \\
            \rowcolor{tGreen}{\fontfamily{ppl}\selectfont \textbf{\#Your Choice\#:} The best answer is Answer 1.} \\
            \rowcolor{tGreen}{\fontfamily{ppl}\selectfont ...} \\
            \rowcolor{tGreen}{\fontfamily{ppl}\selectfont <Demonstrations>} \\
            \specialrule{0em}{1pt}{1pt}
            {\fontfamily{ppl}\selectfont You should try your best to select the best and correct answer. If the two answers are the same, you can randomly choose one. If both answers are incorrect, choose the better one. You MUST select an answer from the provided two answers.} \\
            \\
            {\fontfamily{ppl}\selectfont \textbf{\#Knowledge\#:} <insert the related knowledge>} \\
            {\fontfamily{ppl}\selectfont \textbf{\#Question\#:} <insert the question>} \\
            {\fontfamily{ppl}\selectfont \textbf{\#Answer 1\#:} <insert the hallucinated answer generated by the one-pass schema>} \\
            {\fontfamily{ppl}\selectfont \textbf{\#Answer 2\#:} <insert the hallucinated answer generated by the conversational schema>} \\
            {\fontfamily{ppl}\selectfont \textbf{\#Your Choice\#:}} \\
            \bottomrule
	\end{tabular}
	\caption{Instruction of hallucination filtering for question answering.}
	\label{tab:qa-filter-instruction}
\end{table*}

\begin{table}[t]
\renewcommand\arraystretch{1.1}
\setlength\tabcolsep{3pt}
	\centering
	\small
	\begin{tabular}{r p{0.33\textwidth}}
		\toprule[1pt]
		\textbf{Question} & In what political party was the man who officially opened Royal Spa Centre in 1972? \\
            \midrule[0.5pt]
            \textbf{\tabincell{r}{Right Answer}} & Conservative \\
            \midrule[0.5pt]
            \textbf{\tabincell{r}{Hallucinated\\Answer}} & Labour Party \\
            \midrule[0.5pt]
            \midrule[0.5pt]
            \textbf{User Query} & Retrieve the oldest photo of a cat \\
            \midrule[0.5pt]
            \textbf{ChatGPT} & There is no definitive answer to this question as ``the oldest photo of a cat'' is subjective and varies based on different sources and definitions. However, one candidate for the oldest cat photo is the daguerreotype ``Cat with Blue Ribbon,'' taken by photographer Antoine Claudet in 1840. The photo depicts a cat sitting on a table, gazing directly at the camera, and wearing a blue ribbon tied around its neck. [...] \\
            \midrule[0.5pt]
            \textbf{Hallucination} & Yes \\
            \midrule[0.5pt]
            \textbf{\tabincell{r}{Fragments}} & the oldest cat photo is the daguerreotype ``Cat with Blue Ribbon'' taken by photographer Antoine Claudet in 1840. \\
		\bottomrule[1pt]
	\end{tabular}%
	\caption{A generated hallucinated QA example and a human-labeled ChatGPT response for a user query.}
        \label{tab-benchmark-example}%
        \vspace{-0.3cm}
\end{table}

\subsection{Human Annotation}
\label{sec-annotation}

Besides generating hallucinated samples, we also invite human labelers  to annotate whether ChatGPT responses contain hallucinated content. 

We annotate the general user queries and ChatGPT responses from the $52$K instruction tuning dataset from Alpaca~\cite{alpaca}, which has been widely used by recent LLMs. To screen out user queries where LLMs are most likely to produce hallucination for labeling, we design a pre-selection procedure. Specifically, we use ChatGPT to sample three responses for each user query and compute their average semantic similarity using BERTScore~\cite{ZhangKWWA20}. We finally retain $5,000$ user queries with the lowest similarities. According to recent work~\cite{selfcheckgpt}, hallucinations are likely to appear in diverged and conflicting responses of LLMs. For each query and ChatGPT response, human labelers will annotate whether the response contains hallucinated information (``Yes'' or ``No'') and list the corresponding spans. The hallucination is considered from the following three aspects: \emph{unverifiable}, \emph{non-factual}, and \emph{irrelevant}. Each response is labeled by three human labelers, and we adopt the max-voting strategy to determine the final hallucination label. 

\begin{figure*}[t]
	\centering
	\subfigure[QA Topics]{
		\centering
		\includegraphics[width=0.3\textwidth]{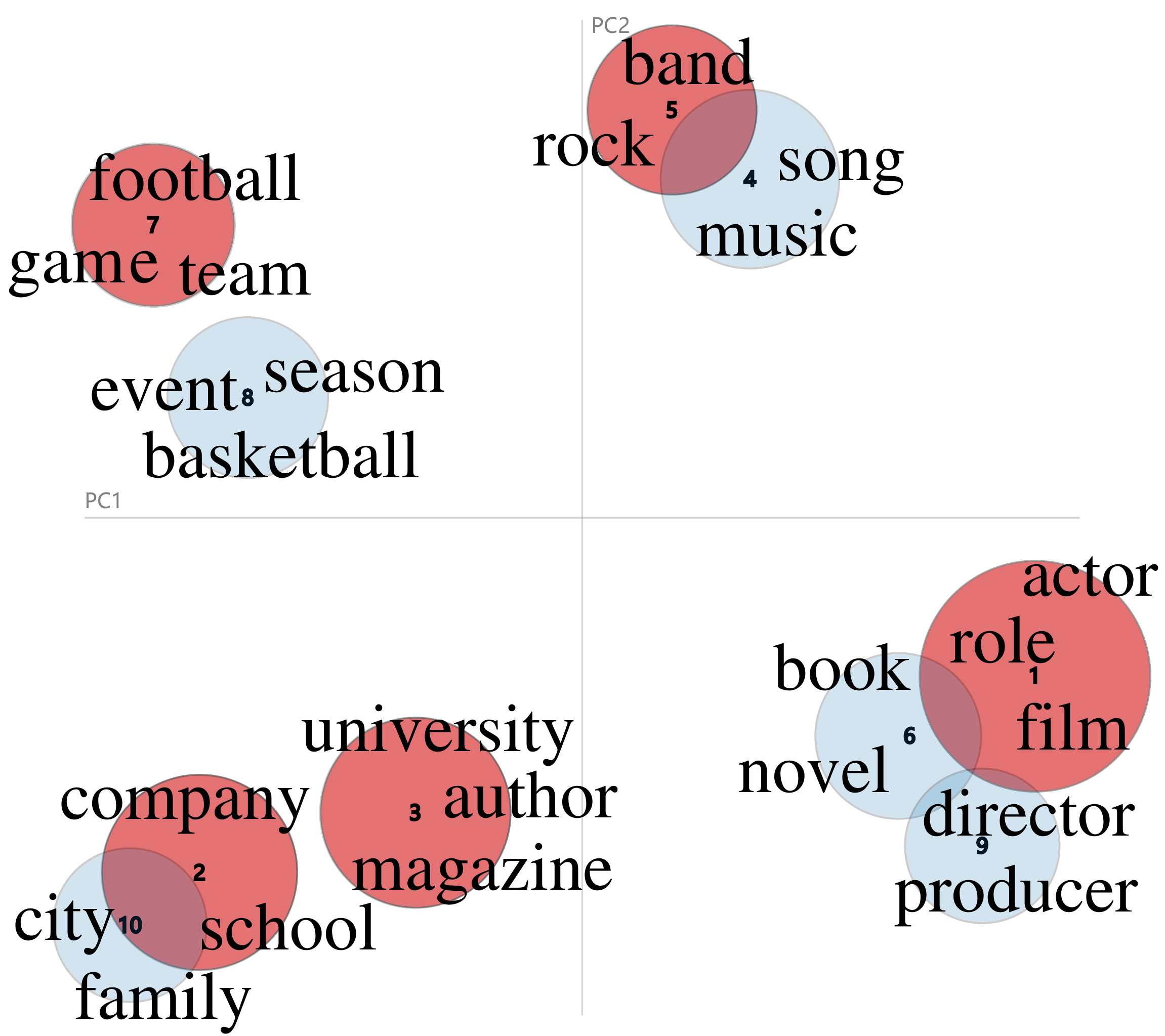}
	}
	\subfigure[Dialogue Topics]{
		\centering
		\includegraphics[width=0.31\textwidth]{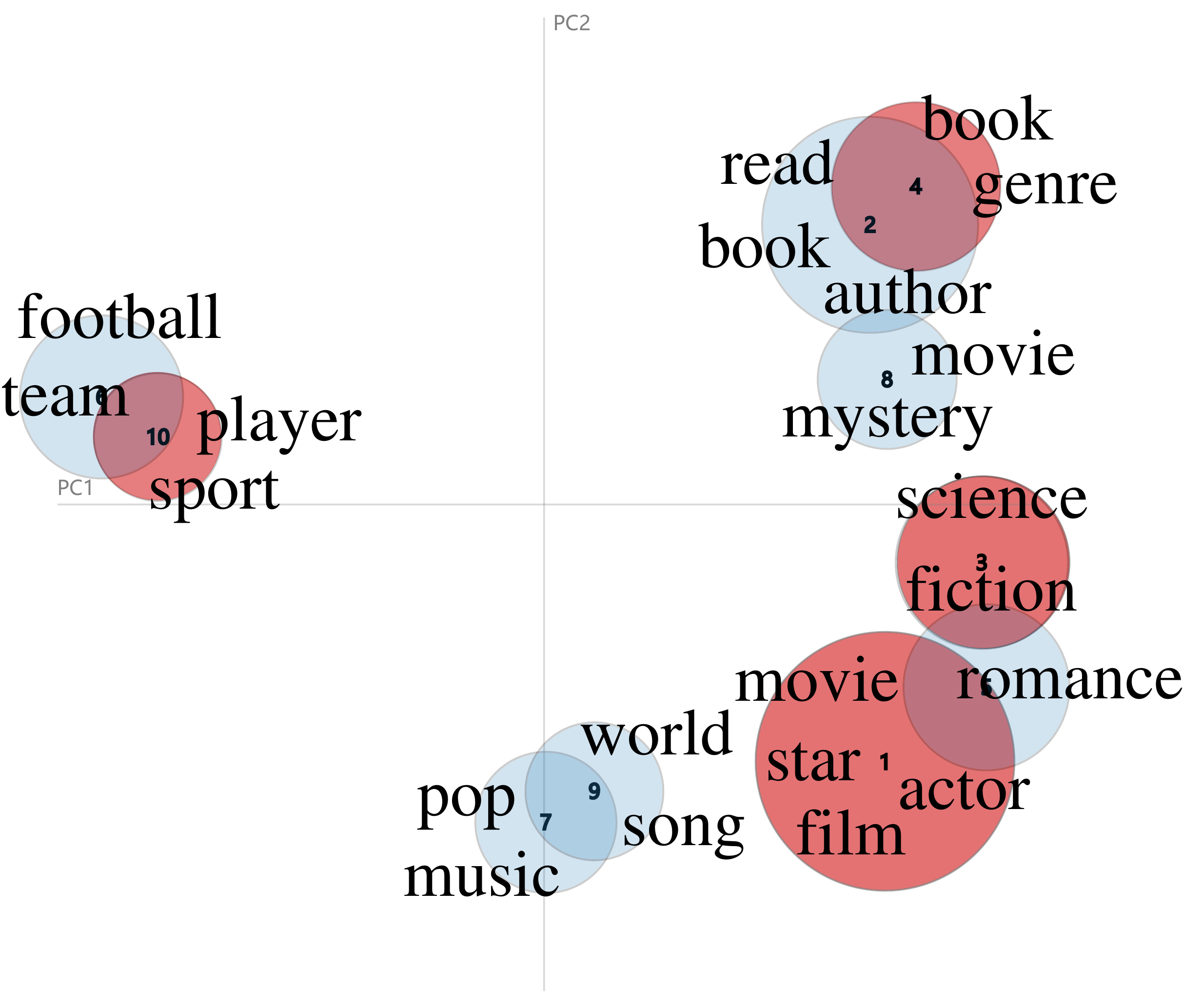}
	}
        \subfigure[Summarization Topics]{
		\centering
		\includegraphics[width=0.31\textwidth]{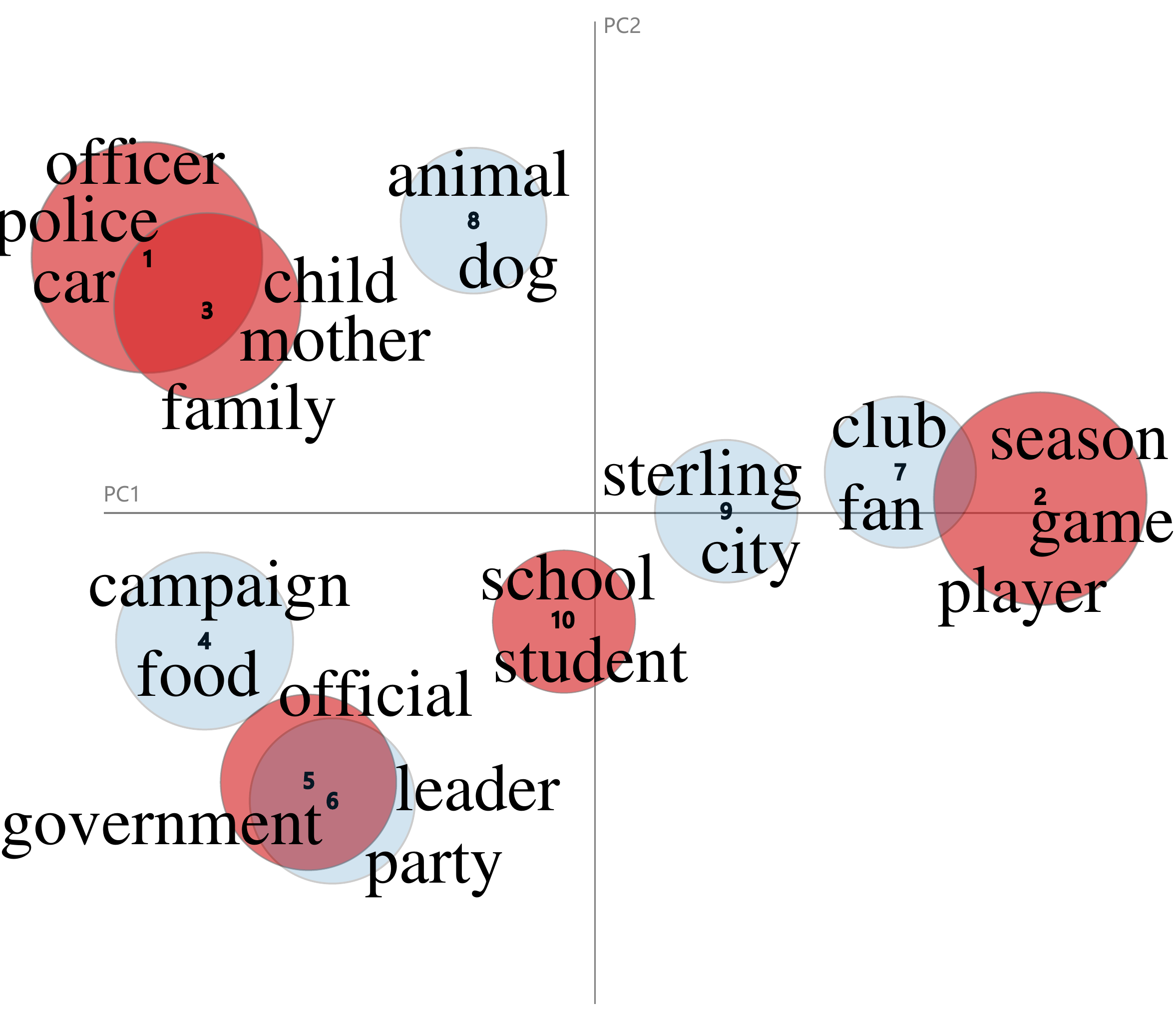}
	}
	\centering
	\caption{Topic distributions for QA, knowledge-grounded dialogue, and text summarization. The samples of each task are classified into 10 topics, and the red circles denote the topics of failed recognized samples by ChatGPT.}
	\label{fig-task-topics}
	\vspace{-0.2cm}
\end{figure*}

\begin{figure}[tb]
	\centering
	\includegraphics[width=0.33\textwidth]{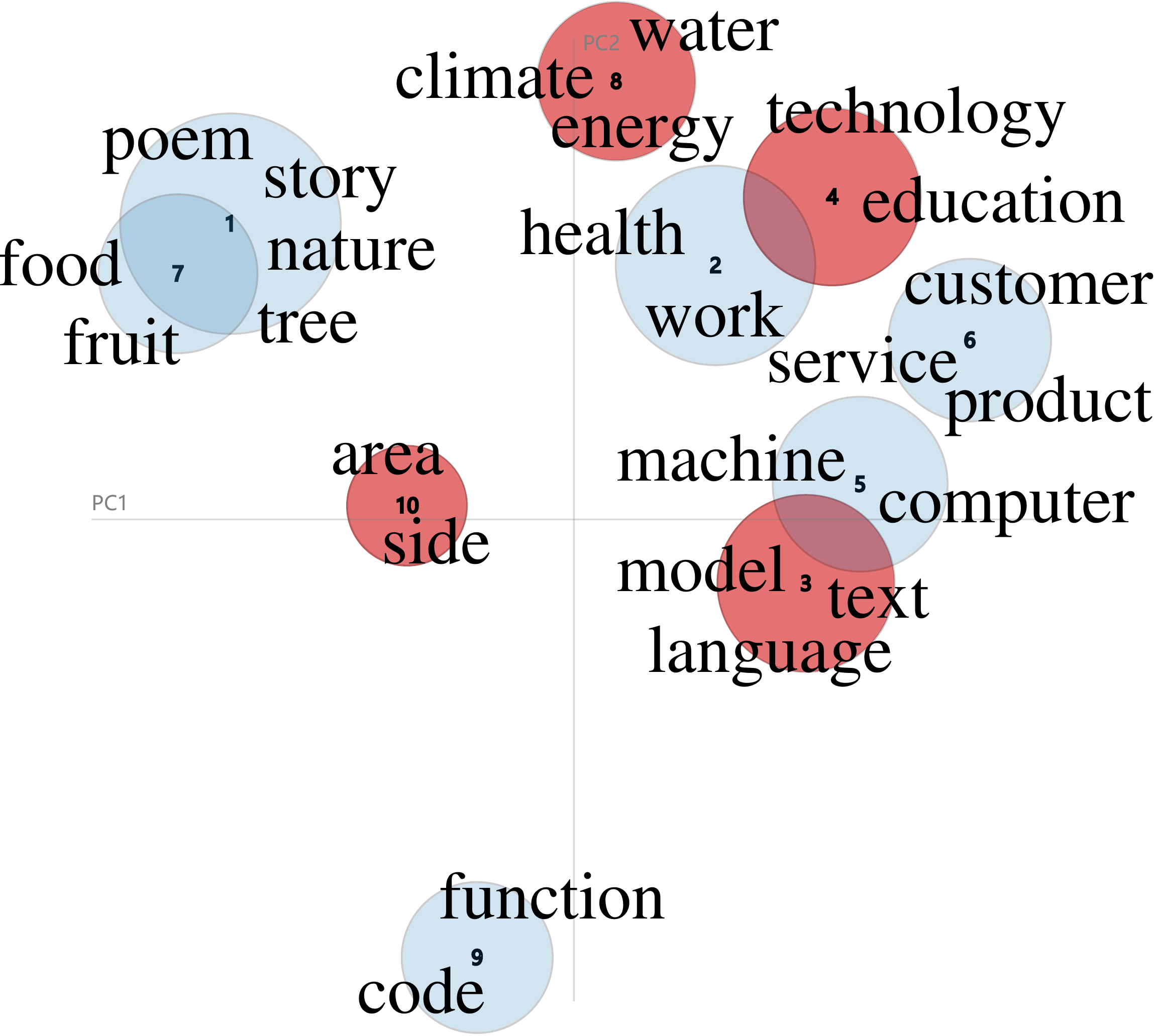}
	\caption{Topic distribution for ChatGPT responses.}
	\label{fig-general-topics}
	\vspace{-0.2cm}
\end{figure}

\paratitle{Labeler Details.} Annotating the hallucination in ChatGPT responses is a very challenging task, which requires good reading comprehension skills and using search engine to look up relevant information for judgement. Thus, from an initial pool of labeler candidates, we select labelers who are good at English passage reading with at least an undergraduate-level education. Besides, following \cite{InstructGPT}, we have labelers annotate a small number of test examples and measure their agreement with the labels of researchers, and finally we choose thirty human labelers with the highest agreement scores. We report Fleiss's Kappa ($\kappa$) to indicate the reliability of agreement between human labelers. We compute $\kappa$ on $5,000$ annotated samples and obtain $\kappa=0.811$ ($0.80 \leq \kappa \leq 1.00$) showing a perfect agreement.

\subsection{Benchmark Analysis and Usage}

With the automatic two-step generation process in Section~\ref{sec-generation}, we produce a total of $30,000$ hallucinated samples with $10,000$ examples for each task of QA, dialogue, and summarization. 
We show the number of generated samples for each hallucination pattern in Table~\ref{tab:overall-hallucination-pattern} at the Appendix~\ref{app-details}. 
Moreover, we manually annotate $5,000$ ChatGPT responses for general user queries in Section~\ref{sec-annotation}. We present a QA example and an annotated query and response example in Table~\ref{tab-benchmark-example}. 
Among the annotated ChatGPT responses, $977$ responses are labeled as containing hallucination ($19.5\%$). Finally, we present the topic distributions of our generated task-specific samples and annotated ChatGPT responses in Figure~\ref{fig-task-topics} and Figure~\ref{fig-general-topics}, ranging from film, sports to school, computer, technology, etc. 

With  our benchmark, researchers can use it to investigate or  mitigate the hallucination issue for LLMs in three aspects. Firstly,  based on our generated and annotated samples, researchers  can use them to analyze what types of content LLMs tend to generate hallucinations. Second, researchers can further evaluate the ability of LLMs to recognize hallucinations in the generated samples. For example, given a question and an answer, LLMs can be asked to determine whether the answer contains hallucinated content. Finally, our benchmark can be further paired with human annotation to assess whether the LLMs' output contains hallucinations, since the samples in our benchmark are specially designed  for testing the hallucinations of LLMs. 

To use our benchmark, users can run the code in our project repository to conduct the corresponding evaluation and analysis. Users can use our provided instructions on their own datasets to evaluate LLMs on hallucinations.

\section{Experiments}

\begin{table}[t]
	\small
	\centering
	\setlength\tabcolsep{3.3pt}
	\begin{tabular}{l c c c c}
		\toprule
		 \textbf{Models} & \textbf{QA} & \textbf{Dialogue} & \textbf{Summarization} & \textbf{General} \\
		\midrule
            \textbf{ChatGPT}  & 62.59   & 72.40   & 58.53 & 79.44 \\
            \textbf{Claude 2}  & 69.78   & 64.73   & 57.75 & 75.00 \\
            \textbf{Claude} & 67.60 & 64.83 & 53.76 & 73.88\\
            \midrule
            \textbf{Davinci002} & 60.05  & 60.81 & 47.77 & 80.42 \\
            \textbf{Davinci003} & 49.65 & 68.37 & 48.07 & 80.40 \\
		\textbf{GPT-3}    & 49.21 &  50.02  &  51.23 & 72.72 \\
		\midrule
            \textbf{Llama 2} & 49.60 & 43.99 & 49.55 & 20.46 \\
            \textbf{ChatGLM} & 47.93 & 44.41 & 48.57 & 30.92 \\
            \textbf{Falcon} & 39.66 & 29.08 & 42.71 & 18.98 \\
            \textbf{Vicuna} & 60.34 & 46.35 & 45.62 & 19.48 \\
            \textbf{Alpaca} & 6.68 & 17.55 & 20.63 & 9.54 \\
		\bottomrule
	\end{tabular}
	\caption{Accuracy (\%) of classifying whether a sample contains hallucinated contents.}
        \vspace{-0.3cm}
	\label{tab:acc-identify}
\end{table}

\subsection{Experimental Setup}

\paratitle{Evaluation Models.} We evaluate several state-of-the-art LLMs in HaluEval benchmark. First, we experiment on five closed-source LLMs, including OpenAI's GPT-3 (\texttt{davinci})~\cite{gpt3}, InstructGPT (\texttt{text-davinci-002/003})~\cite{InstructGPT}, ChatGPT (\texttt{gpt-3.5-turbo}) and Anthropic's \texttt{Claude} and \texttt{Claude 2} models, which can only be accessed through their APIs. Besides, we also evaluate five prevalent open-source LLMs, including \texttt{Alpaca (7B)}~\cite{alpaca}, \texttt{Vicuna (7B)}~\cite{vicuna2023}, \texttt{ChatGLM (7B)}~\cite{Zeng-arxiv-2022-GLM}, \texttt{Falcon (7B)}~\cite{TII-Web-2023-Falcon}, and \texttt{Llama 2-Chat (7B)}~\cite{Touvron-2023-llama2-arxiv}. Our experiments were performed without fine-tuning or engaging in the tuning of hyper-parameters.

\paratitle{Implementation Details.} We execute the generation process of hallucinated samples using Azure OpenAI ChatGPT API. We use a temperature of $1.0$ to generate samples and set the maximum number of tokens for generation to $256$. Moreover, we set the frequency penalty to zero and top-$p$ to $1.0$. For evaluation, we set the temperature to zero for all models to reduce output randomness and ensure more focused and deterministic outputs.

In the following, we first conduct hallucination recognition experiments, then propose several potentially useful strategies to improve the recognition, and finally we perform qualitative analysis to understand the hallucination in LLMs.



\subsection{Results and Analysis}

\subsubsection{Hallucination Recognition}

To evaluate the ability of LLMs to recognize hallucinations, we randomly select the hallucinated or normal output (\eg an answer) of each sample for classification. The evaluation instructions of QA, dialogue, and summarization are presented in Table~\ref{tab:qa-eva-instruction}, Table~\ref{tab:dialogue-eva-instruction} and Table~\ref{tab:summary-eva-instruction} in Appendix~\ref{app-evaluation}.

Table~\ref{tab:acc-identify} presents the accuracy of evaluated LLMs to classify whether the sample output contains hallucinated information. Our findings indicate that LLMs are still poor at identifying hallucination which might be implicit in text. For example, the state-of-the-art ChatGPT model cannot distinguish between factual and hallucinated summary and only achieves 58.53\% accuracy in text summarization, which is barely above chance. Moreover, GPT-3 obtains just about random chance of 50\% accuracy across three tasks, and Alpaca or Vicuna even performs worse (well below random chance). We hypothesize that LLMs perform poorly because the hallucinated sample we generate looks highly similar with ground-truth ones but differs in the key factual spans. As we can see, from GPT-3 to InstructGPT and ChatGPT, instruction tuning and alignment with humans can strength the ability of LLMs in identifying the hallucinations in text.

With respect to the hallucinated samples where ChatGPT fails to recognize, we present the number of each hallucination pattern in Table~\ref{tab:hallucination-pattern}. 
Based on the results, we can observe that the hallucination patterns of failed samples are unevenly distributed. For example, over half of failures in QA, dialogue, and summarization originate from the first hallucination pattern (\ie \emph{comprehension}, \emph{extrinsic-soft}, and \emph{factual}), which refers to the hallucinations that are factually correct but conflict with the context. This indicates that LLMs lack or cannot associate related knowledge to identify the factual hallucination in the generated text. 
To further understand the failures of ChatGPT, we visualize the topics of those failed samples via Latent Dirichlet Allocation (LDA)~\cite{lda}. As shown in Figure~\ref{fig-task-topics} and Figure~\ref{fig-general-topics}, we cluster all task samples into ten topics and mark the topics of failed samples as red. We find that the hallucination of LLMs is topic-sensitive. For example, the frequent topics in QA include film, school, and company. While, ChatGPT mainly fails to recognize those samples in the topics of film, company, and band. For user queries and ChatGPT responses, the top five topics include story, health, language, technology, and computer. ChatGPT mainly faces challenges in topics of technology, climate, and language.

\begin{table}[t]
	\small
	\centering
	\setlength\tabcolsep{4pt}
	\begin{tabular}{l c c c c c}
		\toprule
		  \textbf{Tasks} & \textbf{\#Failed} & \textbf{P-I} & \textbf{P-II} & \textbf{P-III} & \textbf{P-IV} \\
		\midrule
		\textbf{QA} & 3109  &  1559 & 245 & 278 & 1027 \\
            \textbf{Dialogue}  & 891  & 465  & 344 & 82 & - \\
            \textbf{Summarization}  & 3868 & 3106  & 705 & 57 & - \\
		\bottomrule
	\end{tabular}
	\caption{Number of samples where ChatGPT fails to recognize for each hallucination pattern (P-I/II/III/IV). }
        \vspace{-0.3cm}
	\label{tab:hallucination-pattern}
\end{table}

\begin{table*}[t]
\renewcommand\arraystretch{1.1}
\setlength\tabcolsep{3pt}
	\centering
	\small
	\begin{tabular}{c | p{0.43\textwidth} | p{0.43\textwidth}}
		\toprule[1pt]
		\makecell*[c]{\textbf{User Query}} & \makecell*[l]{Generate a list of 5 important dates in US history.} & \makecell*[l]{Create a visualization to compare the GDP growth of \\India and China between 1998 and 1998.} \\
            \hline
            \textbf{\makecell*[c]{ChatGPT\\Response}} & \tabincell{l}{\textcolor{fGreen}{\textbf{1. July 4, 1776 - Declaration of Independence}} \textcolor{fGreen}{\textbf{sign-}}\\\textcolor{fGreen}{\textbf{ing}}\\
            2. April 12, 1861 - Start of the Civil War\\
            3. August 18, 1920 - 19th Amendment ratification gr-\\anting women the right to vote\\
            4. December 7, 1941 - Pearl Harbor attack during Wo-\\rld War II\\
            5. September 11, 2001 - Terrorist attacks on New York\\ City and Washington D.C.}
 & \makecell*[l]{As an AI language model, I cannot create visualizations,\\ but I can provide you with the following information: \\According to World Bank data, India and China have\\ shown significant economic growth in the past two de-\\cades. \textcolor{fGreen}{\textbf{In 1998, India's GDP growth rate was 4.8\%,}}\\\textcolor{fGreen}{\textbf{while China's GDP growth rate was 9.3\%.}}\\ In the following years, the GDP growth rate fluctuat-\\ed, but China maintained a higher growth rate than \\India. \textcolor{fGreen}{\textbf{By 2019, China's GDP growth rate was 6.1\%, }}\\\textcolor{fGreen}{\textbf{while India's GDP growth rate was 4.2\%.}}} \\
            \hline
            \textbf{\makecell*[c]{Retrieved\\Knowledge}} & \makecell*[l]{The signing of the United States Declaration of Inde-\\pendence occurred primarily on August 2, 1776, at the\\ Pennsylvania State House in Philadelphia, later known \\as Independence Hall.\\\url{https://en.wikipedia.org/wiki/Signing_of_the}\\\url{_United_States_Declaration_of_Independence}} & \makecell*[l]{India’s GDP growth rate in 1998 was 6.2\%. China’s \\GDP growth rate in 1998 was 7.8\%. \\India’s GDP growth rate in 2019 was 3.7\%. China’s \\GDP growth rate in 2019 was 6.0\%.\\\url{https://data.worldbank.org/}}\\
            \hline
            \textbf{\makecell*[c]{Refined\\Response}} & \tabincell{l}{\textcolor{gold}{\textbf{1. August 2, 1776 - Declaration of Independence si-}}\\\textcolor{gold}{\textbf{gning}}\\
            2. April 12, 1861 - Start of the Civil War\\
            3. August 18, 1920 - 19th Amendment ratification gr-\\anting women the right to vote\\
            4. December 7, 1941 - Pearl Harbor attack during Wo-\\rld War II\\
            5. September 11, 2001 - Terrorist attacks on New York\\ City and Washington D.C.} & \makecell*[l]{According to World Bank data, India and China have\\ shown significant economic growth in the past two de-\\cades. \textcolor{gold}{\textbf{In 1998, India's GDP growth rate was 6.2\%,}}\\\textcolor{gold}{\textbf{while China's GDP growth rate was 7.8\%. By 2019, }}\\\textcolor{gold}{\textbf{China's GDP growth rate was 6.0\%, while India's}}\\\textcolor{gold}{\textbf{GDP growth rate was 3.7\%.}}} \\
		\bottomrule[1pt]
	\end{tabular}%
	\caption{Two hallucinated and refined examples from ChatGPT. The \textcolor{fGreen}{\textbf{green}} text denotes the hallucinated span, and the \textcolor{gold}{\textbf{brown}} text denotes the refined span after augmented with retrieved knowledge.}
        \label{tab-case-study}%
        \vspace{-0.3cm}
\end{table*}

\begin{table}[t]
	\small
	\centering
	\setlength\tabcolsep{2.0pt}
	\begin{tabular}{l c c c c}
		\toprule
		  \textbf{Variants} & \textbf{QA} & \textbf{Dialogue} & \textbf{Summarization} & \textbf{General} \\
		\midrule
		\textbf{ChatGPT} & 62.59  & 72.40 & 58.53 & 86.22 \\
            \midrule
		w/ Knowledge & 76.83 & 73.80 & - & 90.73 \\
            w/ CoT & 59.58 & 71.39 & 61.21 & 86.50 \\
            w/ Contrast & 49.19 & 68.67 & 49.46 & - \\
		\bottomrule
	\end{tabular}
	\caption{Accuracy (\%) of ChatGPT equipped with three improvement strategies.}
        \vspace{-0.3cm}
	\label{tab:improvement}
\end{table}

\subsubsection{Improvement Strategies}

In this part, we design several strategies to improve the ability of LLMs to recognize hallucination. The results are shown in Table~\ref{tab:improvement}.

\paratitle{Knowledge Retrieval.} Retrieving relevant knowledge is a widely used strategy to eliminate hallucination~\cite{rag,li2023web}. Therefore, we supply ChatGPT with the knowledge facts retrieved from Wikipedia (except for that summarization does not need external information besides the source document).  By providing knowledge, the recognition accuracy of ChatGPT increases significantly (\eg increasing from 62.59 to 76.83 in QA), while the performance improvement in dialogue is mild. We hypothesize that the common hallucination patterns in dialogue (\ie extrinsic-soft/hard) cannot be simply identified via incorporating external knowledge. For those general user queries and ChatGPT responses, we discover that providing external knowledge does have a significant benefit. 
Thus, equipping LLMs with external knowledge can largely enhance their abilities to recognize hallucinations.

\paratitle{CoT Reasoning.} In previous work~\cite{cot}, chain-of-thought (CoT) has been proposed to improve the ability of LLMs to perform reasoning and derive the final answer by introducing a series of intermediate reasoning steps. Here, besides producing the recognition result, we also require ChatGPT to generate the reasoning steps. While, from the results in Table~\ref{tab:improvement}, we observe that generating reasoning steps can mildly improve the performance but makes the model perform worse in QA and dialogue (\eg dropping from 62.59 to 59.58). Compared to retrieving knowledge, adding chain-of-thought before output might interfere with the final judgement. While, in text summarization, generating reasoning steps improve the accuracy from 58.53 to 61.21. The reason might be that the factual contradiction between document and summary can be identified through logic reasoning.

\paratitle{Sample Contrast.} We further provide ground-truth examples for ChatGPT to test whether it can distinguish the right sample from the hallucinated sample. As we can see from Table~\ref{tab:improvement}, distinguishing between right and hallucinated samples achieves the worst results. We hypothesize that our generated hallucinated samples have a high similarity to the real samples, thus making LLMs confused to distinguish them. This test also indicates that our benchmark is very challenging in hallucination evaluation for LLMs.

\subsection{Case Study}

In the above, we have observed that providing external knowledge can be  beneficial for LLMs to mitigate and recognize hallucinations. To demonstrate the effectiveness of knowledge retrieval in mitigating hallucinations, we present two hallucinated responses from ChatGPT and refined responses after augmented with retrieved knowledge in Table~\ref{tab-case-study}. In the first example, the generated span (\ie ``{July 4, 1776 - Declaration of Independence signing}'') contains hallucinated information because it gives a wrong time of Declaration of Independence signing. By providing retrieved information about Declaration of Independence signing, ChatGPT is able to correct the hallucinated span and give the right information. Analogously, in the second example, ChatGPT gives incorrect GDP growth rates of China and India, which is due to that API-based ChatGPT cannot access the web to obtain the official data. After providing official information retrieved from World Bank, the refined span displays answers that contain the correct information. The above two examples illustrate that retrieving knowledge related to queries can help ChatGPT significantly reduce the hallucinations in the response, especially those factual errors.
\section{Related Work}

\paratitle{Hallucination in LLMs.} Hallucination in LLMs is concerning since it hinders performance and raises safety risks in real-world application. To alleviate this issue, prior studies have proposed to use a verification system to identify non-factual entities in text summarization~\cite{Zheng-Reducing}, invoke interfaces of structured data (\eg knowledge graph, database) to obtain related evidence~\cite{structGPT,lan2022complex}, and train a token-level fact critic to recognize hallucination and rectify them in dialogue~\cite{DziriMZB21}. To enhance the understanding of hallucination in LLMs and promote the unification of research efforts, there are many active endeavors to analyze the causes of hallucination in different tasks and investigate their relationship~\cite{QA-Hallucination,dialogue-hallucination,summarization-hallucination}. Our work is closely related to these work, but we focus on building a hallucination evaluation benchmark for LLMs. Our dataset can serve as a public platform for exhibiting the blind spots of LLMs in solving hallucination.

\paratitle{Hallucination Evaluation.} Another line of work focusing on evaluating the hallucination of models in different NLP tasks~\cite{DziriRLR22,GuptaWLX22,DziriKMZYPR22,Rashkin-Measuring,Li-hallucination-2023}. For instance, The BEGIN benchmark~\cite{DziriRLR22} classifies the utterances generated by dialogue systems into three categories, \ie fully attributable, not fully attributable, and generic; and the Attributable to Identified Sources (AIS) benchmark~\cite{Rashkin-Measuring} assesses whether the source documents support the output of text generation models. Though these benchmarks can serve as decent evaluation platform, they are penurious in only focusing on single tasks (\eg dialogue) and small models (\eg DPR). Besides, several metrics have been proposed to quantify hallucination, such as PARENT~\cite{DhingraFPCDC19} for measuring $n$-gram lexical entailment in table-to-text generation and TRUE~\cite{HonovichAHTKCSS22} computes the example-level Area Under the ROC Curve. In this work, our HaluEval benchmark includes general user queries and ChatGPT responses and proposes a two-step automatic process to generate hallucinated samples for evaluation, which is completely based on LLMs. 
\section{Conclusion}

We introduce HaluEval, a large-scale collection of generated and human-annotated hallucinated samples for evaluating the performance of LLMs in recognizing hallucinations. To automatically generate large-scale samples, we propose a two-step approach, \ie sampling-then-filtering. We first introduce two different sampling methods to generate diverse samples using instructions and then filter and select the difficult one. Besides, we invite qualified human labelers to annotate the hallucinations of ChatGPT responses given user queries. We find that, existing LLMs mostly fail to recognize the hallucinations in text and tend to generate hallucinated content. Finally, we suggest several strategies to help LLMs recognize hallucinations. Our benchmark can facilitate research in understanding what types of content and to which extent LLMs tend to hallucinate, ultimately paving the way for building more effective and reliable LLMs in the future.
\section{Limitations}

In our approach, we leverage a LLM, \ie ChatGPT, to automatically generate the hallucinated samples. Therefore, the quality of our hallucinated samples is limited by the capacity of ChatGPT in following the complex instruction of hallucination sampling. 
Although we design the high-quality hallucination filtering process, it is still necessary to apply quality control to the generation of hallucinated samples.
Besides, our benchmark focuses on evaluating the ability of LLMs in recognizing the hallucinations in text but does not investigate the underlying reasons behind the appearance of hallucinations like prior work~\cite{QA-Hallucination,dialogue-hallucination}. 

As for the potential issue, since the hallucinated samples in our benchmark looks highly similar to the ground-truth samples, which might be misused for an unexpected purpose than we planned. To alleviate this issue, we should monitor and regulate the spread and usage of our benchmark.

\section*{Acknowledgments}

This work was partially supported by National Natural Science Foundation of China under Grant No. 62222215, Beijing Natural Science Foundation under Grant No. L233008 and 4222027, and Beijing Outstanding Young Scientist Program under Grant No. BJJWZYJH012019100020098. And this work is also partially supported by the Outstanding Innovative Talents Cultivation Funded Programs 2021 of Renmin University of China. Xin Zhao is the corresponding author.

\bibliography{anthology}
\bibliographystyle{acl_natbib}

\newpage
\appendix
\section*{Appendix}

We provide some extra information about our benchmark as supplementary materials. The appendix is organized into three sections:
\begin{itemize}
	\item Instructions of hallucination sampling are presented in Appendix~\ref{app-sampling};
	\item Instructions of hallucination filtering are presented in Appendix~\ref{app-filtering};
        \item Instructions of evaluation are presented in Appendix~\ref{app-evaluation};
	\item Details of our benchmark are presented in Appendix~\ref{app-details}.
\end{itemize}

\section{Hallucination Sampling} \label{app-sampling}

The hallucination sampling instructions for dialogue and summarization are shown in Table~\ref{tab:dialogue-gen-instruction} and Table~\ref{tab:summary-gen-instruction}, respectively.

\begin{table*}[t]
	\small
	\centering
	\begin{tabular}{p{0.97\textwidth}}
		\toprule
            \rowcolor{LightSkyBlue1}{\fontfamily{ppl}\selectfont I want you act as an assistant in a conversation with human. Given a dialogue history, the true response, and related knowledge, your objective is to write a hallucinated response that sounds plausible but is factually incorrect. You SHOULD write the hallucinated response using the following method (each with some examples):} \\
            \\
            \rowcolor{tPink}{\fontfamily{ppl}\selectfont You are trying to write a response to human but you replace the true entity with a highly similar entity.} \\
            \specialrule{0em}{1pt}{1pt}
            \rowcolor{tGreen}{\fontfamily{ppl}\selectfont \textbf{\#Knowledge\#:} The Dark Knight is a 2008 superhero film directed by Christopher Nolan from a screenplay he co-wrote with his brother Jonathan. Christopher Nolan is a film director.} \\
            \rowcolor{tGreen}{\fontfamily{ppl}\selectfont \textbf{\#Dialogue History\#:} [Human]: Could you recommend movies similar to The Dark Knight? [Assistant]: The sequel to Batman Begins is The Dark Knight. [Human]: Okay. Who is the director of The Dark Knight and any other movies from him not related to Batman?} \\
            \rowcolor{tGreen}{\fontfamily{ppl}\selectfont \textbf{\#True Response\#:} Christopher Nolan was the director. He also directed insomnia and inception.} \\
            \rowcolor{tGreen}{\fontfamily{ppl}\selectfont \textbf{\#Hallucinated Response\#:} Steven Spielberg was the director. He also directed insomnia and inception.} \\
            {\fontfamily{ppl}\selectfont or} \\
            \rowcolor{tPink}{\fontfamily{ppl}\selectfont You are trying to write a response to human but you replace the true entity with a dissimilar entity.} \\
            \specialrule{0em}{1pt}{1pt}
            \rowcolor{tGreen}{\fontfamily{ppl}\selectfont <Demonstrations>} \\
            {\fontfamily{ppl}\selectfont or} \\
            \rowcolor{tPink}{\fontfamily{ppl}\selectfont You are trying to write a response to human but you replace the true entity with a dissimilar entity in a different entity type.} \\
            \specialrule{0em}{1pt}{1pt}
            \rowcolor{tGreen}{\fontfamily{ppl}\selectfont <Demonstrations>} \\
            \\
            {\fontfamily{ppl}\selectfont You should try your best to make the response become hallucinated. } \\
            \\
            {\fontfamily{ppl}\selectfont \textbf{\#Knowledge\#:} <Here is the related knowledge>} \\
           {\fontfamily{ppl}\selectfont \textbf{\#Dialogue History\#:} <Here is the dialogue history>} \\
            {\fontfamily{ppl}\selectfont \textbf{\#True Response\#:} <Here is the true response of the dialogue history>} \\
            {\fontfamily{ppl}\selectfont \textbf{\#Hallucinated Response\#:}} \\
            \bottomrule
	\end{tabular}
	\caption{Instruction of hallucination sampling for knowledge-grounded dialogue. } 
	\label{tab:dialogue-gen-instruction}
\end{table*}

\begin{table*}[t]
	\small
	\centering
	\begin{tabular}{p{0.97\textwidth}}
		\toprule
            \rowcolor{LightSkyBlue1}{\fontfamily{ppl}\selectfont I want you act as a hallucination summary generator. Given a document and the right summary, your objective is to write a hallucinated summary that sounds plausible but is factually incorrect. You SHOULD write the hallucinated summary using the following method (each with some examples):} \\
            \\
            \rowcolor{tPink}{\fontfamily{ppl}\selectfont You are trying to write a summary which is factual but some information cannot be directly inferred or entailed from the document.} \\
            \specialrule{0em}{1pt}{1pt}
            \rowcolor{tGreen}{\fontfamily{ppl}\selectfont \textbf{\#Document\#:} The panther chameleon was found on Monday by a dog walker in the wooded area at Marl Park. It had to be put down after X-rays showed all of its legs were broken and it had a deformed spine. RSPCA Cymru said it was an "extremely sad example of an abandoned and neglected exotic pet". Inspector Selina Chan said: "It is a possibility that the owners took on this animal but were unable to provide the care he needs and decided to release him to the wild. "We are urging potential owners of exotic animals to thoroughly research what is required in the care of the particular species before taking one on. "Potential owners need to make sure they can give their animal the environment it needs and they have the facilities, time, financial means and long-term commitment to maintain a good standard of care, as required under the Animal Welfare Act 2006." She added it was illegal to release non-native species into the wild.} \\
            \rowcolor{tGreen}{\fontfamily{ppl}\selectfont \textbf{\#Right Summary\#:} Owners of exotic animals have been urged to do research before having them as pets after a seriously neglected chameleon was found in Cardiff Bay. } \\
            \rowcolor{tGreen}{\fontfamily{ppl}\selectfont \textbf{\#Hallucinated Summary\#:} A chameleon that was found in a Cardiff park has been put down after being abandoned and neglected by its owners.} \\
            {\fontfamily{ppl}\selectfont or} \\
            \rowcolor{tPink}{\fontfamily{ppl}\selectfont You are trying to write a summary but there exist some non-factual and incorrect information. You can fabricate some information that does not exist in the provided document.} \\
            \specialrule{0em}{1pt}{1pt}
            \rowcolor{tGreen}{\fontfamily{ppl}\selectfont <Demonstrations>} \\
            {\fontfamily{ppl}\selectfont or} \\
            \rowcolor{tPink}{\fontfamily{ppl}\selectfont You are trying to write a summary but there is a factual contradiction between the summary and the document.} \\
            \specialrule{0em}{1pt}{1pt}
            \rowcolor{tGreen}{\fontfamily{ppl}\selectfont <Demonstrations>} \\
            \\
            {\fontfamily{ppl}\selectfont You should try your best to make the summary become hallucinated. \#Hallucinated Summary\# can only have about 5 more words than \#Right Summary\#.} \\
            \\
            {\fontfamily{ppl}\selectfont \textbf{\#Document\#:} <Here is the test document>} \\
            {\fontfamily{ppl}\selectfont \textbf{\#Right Summary\#:} <Here is the right summary of the test document>} \\
            {\fontfamily{ppl}\selectfont \textbf{\#Hallucinated Summary\#:}} \\
            \bottomrule
	\end{tabular}
	\caption{Instruction of hallucination sampling for text summarization. } 
	\label{tab:summary-gen-instruction}
\end{table*}

\section{Hallucination Filtering} \label{app-filtering}

The hallucination sampling instructions for dialogue and summarization are shown in Table~\ref{tab:dialogue-filter-instruction} and Table~\ref{tab:summary-filter-instruction}, respectively.

\begin{table*}[t]
	\small
	\centering
	\begin{tabular}{p{0.97\textwidth}}
		\toprule
            \rowcolor{LightSkyBlue1}{\fontfamily{ppl}\selectfont I want you act as a response judge. Given a dialogue history, two responses, and related knowledge, your objective is to select the best and correct response without hallucination and non-factual information. Here are some examples:} \\
            \\
            \rowcolor{tGreen}{\fontfamily{ppl}\selectfont \textbf{\#Knowledge\#:}The Dark Knight is a 2008 superhero film directed by Christopher Nolan from a screenplay he co-wrote with his brother Jonathan. Christopher Nolan is a film director.} \\
            \rowcolor{tGreen}{\fontfamily{ppl}\selectfont \textbf{\#Dialogue History\#:} [Human]: Could you recommand movies similar to The Dark Knight? [Assistant]: The sequel to Batman Begins is The Dark Knight. [Human]: Okay. Who is the director of The Dark Knight and any other movies from him not related to Batman?} \\
            \rowcolor{tGreen}{\fontfamily{ppl}\selectfont \textbf{\#Response 1\#:} Christopher Nolan was the director. He also directed insomnia and inception.} \\
            \rowcolor{tGreen}{\fontfamily{ppl}\selectfont \textbf{\#Response 2\#:} Steven Spielberg was the director. He also directed insomnia and inception.} \\
            \rowcolor{tGreen}{\fontfamily{ppl}\selectfont \textbf{\#Your Choice\#:} The best response is Response 1.} \\
            {\fontfamily{ppl}\selectfont ...} \\
            {\fontfamily{ppl}\selectfont <Demonstrations>} \\
            {\fontfamily{ppl}\selectfont ...} \\
            \\
            {\fontfamily{ppl}\selectfont You should try your best to select the best and correct response. If the two responses are the same, you can randomly choose one. If both responses are incorrect, choose the better one. You MUST select a response from the provided two responses.} \\
            \\
            {\fontfamily{ppl}\selectfont \textbf{\#Knowledge\#:} <Here is the related knowledge>} \\
            {\fontfamily{ppl}\selectfont \textbf{\#Dialogue History\#:} <Here is the dialogue history>} \\
            {\fontfamily{ppl}\selectfont \textbf{\#Response 1\#:} <Here is the hallucinated response generated by the first channel>} \\
            {\fontfamily{ppl}\selectfont \textbf{\#Response 2\#:} <Here is the hallucinated response generated by the second channel>} \\
            {\fontfamily{ppl}\selectfont \textbf{\#Your Choice\#:}} \\
            \bottomrule
	\end{tabular}
	\caption{Instruction of hallucination filtering for knowledge-grounded dialogue.}
        \vspace{-0.2cm}
	\label{tab:dialogue-filter-instruction}
\end{table*}

\begin{table*}[t]
	\small
	\centering
	\begin{tabular}{p{0.97\textwidth}}
		\toprule
            \rowcolor{LightSkyBlue1}{\fontfamily{ppl}\selectfont I want you act as a summary judge. Given a document and two summaries, your objective is to select the best and correct summary without hallucination and non-factual information. Here are some examples:} \\
            \\
            \rowcolor{tGreen}{\fontfamily{ppl}\selectfont \textbf{\#Document\#:}The panther chameleon was found on Monday by a dog walker in the wooded area at Marl Park. It had to be put down after X-rays showed all of its legs were broken and it had a deformed spine. RSPCA Cymru said it was an "extremely sad example of an abandoned and neglected exotic pet". Inspector Selina Chan said: "It is a possibility that the owners took on this animal but were unable to provide the care he needs and decided to release him to the wild. "We are urging potential owners of exotic animals to thoroughly research what is required in the care of the particular species before taking one on. "Potential owners need to make sure they can give their animal the environment it needs and they have the facilities, time, financial means and long-term commitment to maintain a good standard of care, as required under the Animal Welfare Act 2006." She added it was illegal to release non-native species into the wild.} \\
            \rowcolor{tGreen}{\fontfamily{ppl}\selectfont \textbf{\#Summary 1\#:} Owners of exotic animals have been urged to do research before having them as pets after a seriously neglected chameleon was found in Cardiff Bay.} \\
            \rowcolor{tGreen}{\fontfamily{ppl}\selectfont \textbf{\#Summary 2\#:} A chameleon that was found in a Cardiff park has been put down after being abandoned and neglected by its owners.} \\
            \rowcolor{tGreen}{\fontfamily{ppl}\selectfont \textbf{\#Your Choice\#:} The best summary is Summary 1.} \\
            {\fontfamily{ppl}\selectfont ...} \\
            {\fontfamily{ppl}\selectfont <Demonstrations>} \\
            {\fontfamily{ppl}\selectfont ...} \\
            \\
            {\fontfamily{ppl}\selectfont You should try your best to select the best and correct summary. If both summaries are incorrect, choose the better one. You MUST select a summary from the provided two summaries.} \\
            \\
            {\fontfamily{ppl}\selectfont \textbf{\#Document\#:} <Here is the test document>} \\
            {\fontfamily{ppl}\selectfont \textbf{\#Summary 1\#:} <Here is the hallucinated summary generated by the first channel>} \\
            {\fontfamily{ppl}\selectfont \textbf{\#Summary 2\#:} <Here is the hallucinated summary generated by the second channel>} \\
            {\fontfamily{ppl}\selectfont \textbf{\#Your Choice\#:}} \\
            \bottomrule
	\end{tabular}
	\caption{Instruction of hallucination filtering for text summarization.}
        \vspace{-0.2cm}
	\label{tab:summary-filter-instruction}
\end{table*}

\section{Hallucination Recognition} \label{app-evaluation}

The hallucination recognition instructions for QA, dialogue and summarization are shown in Table~\ref{tab:qa-eva-instruction}, Table~\ref{tab:dialogue-eva-instruction} and Table~\ref{tab:summary-eva-instruction}, respectively.

\begin{table*}[t]
	\small
	\centering
	\begin{tabular}{p{0.97\textwidth}}
		\toprule
            \rowcolor{LightSkyBlue1}{\fontfamily{ppl}\selectfont I want you act as an answer judge. Given a question and an answer, your objective is to determine if the provided answer contains non-factual or hallucinated information. You SHOULD give your judgement based on the following hallucination types and the world knowledge.} \\
            \\
            \rowcolor{tPink}{\fontfamily{ppl}\selectfont You are trying to determine if there is a factual contradiction between the answer and the world knowledge. Some information in the answer might be fabricated.} \\
            \specialrule{0em}{1pt}{1pt}
            \rowcolor{tGreen}{\fontfamily{ppl}\selectfont \textbf{\#Question\#:} What U.S Highway gives access to Zilpo Road, and is also known as Midland Trail?} \\
            \rowcolor{tGreen}{\fontfamily{ppl}\selectfont \textbf{\#Answer\#:} U.S. Highway 60} \\
            \rowcolor{tGreen}{\fontfamily{ppl}\selectfont \textbf{\#Your Judgement\#:} No} \\
            \rowcolor{tGreen}{\fontfamily{ppl}\selectfont \textbf{\#Question\#:} Are the New Orleans Outfall Canals the same length as the Augusta Canal?} \\
            \rowcolor{tGreen}{\fontfamily{ppl}\selectfont \textbf{\#Answer\#:} No. The Orleans Canal is approximately 3.6 miles (5.8 kilometers) long while the Augusta Canal is approximately 7 miles (11.3 kilometers) long.} \\
            \rowcolor{tGreen}{\fontfamily{ppl}\selectfont \textbf{\#Your Judgement\#:} Yes} \\
            \\
            \rowcolor{tPink}{\fontfamily{ppl}\selectfont You are trying to determine if the answer misunderstands the question context and intention.} \\
            \specialrule{0em}{1pt}{1pt}
            \rowcolor{tGreen}{\fontfamily{ppl}\selectfont <Demonstrations>} \\
            \\
            \rowcolor{tPink}{\fontfamily{ppl}\selectfont You are trying to determine if the answer is too general or too specific to answer the question at an appropriate level of specificity.} \\
            \specialrule{0em}{1pt}{1pt}
            \rowcolor{tGreen}{\fontfamily{ppl}\selectfont <Demonstrations>} \\
            \\
            \rowcolor{tPink}{\fontfamily{ppl}\selectfont You are trying to determine if the answer cannot be inferred from the knowledge correctly.} \\
            \specialrule{0em}{1pt}{1pt}
            \rowcolor{tGreen}{\fontfamily{ppl}\selectfont <Demonstrations>} \\
            \\
            {\fontfamily{ppl}\selectfont You should try your best to determine if the answer contains non-factual or hallucinated information according to the above hallucination types. The answer you give MUST be ``Yes'' or ``No''.} \\
            \\
            {\fontfamily{ppl}\selectfont \textbf{\#Question\#:} <Here is the test question>} \\
            {\fontfamily{ppl}\selectfont \textbf{\#Answer\#:} <Here is the hallucinated answer or right answer>} \\
            {\fontfamily{ppl}\selectfont \textbf{\#Your Judgement\#:}} \\
            \bottomrule
	\end{tabular}
	\caption{Instruction of hallucination recognition for question answering.}
        \vspace{-0.2cm}
	\label{tab:qa-eva-instruction}
\end{table*}

\begin{table*}[t]
	\small
	\centering
	\begin{tabular}{p{0.97\textwidth}}
		\toprule
            \rowcolor{LightSkyBlue1}{\fontfamily{ppl}\selectfont I want you act as a response judge. Given a dialogue history and a response, your objective is to determine if the provided response contains non-factual or hallucinated information. You SHOULD give your judgement based on the following hallucination types and the world knowledge.} \\
            \\
            \rowcolor{tPink}{\fontfamily{ppl}\selectfont You are trying to determine if the true entity in the response is replaced with a highly similar entity.} \\
            \specialrule{0em}{1pt}{1pt}
            \rowcolor{tGreen}{\fontfamily{ppl}\selectfont \textbf{\#Dialogue History\#:} [Human]: Could you recommend movies similar to The Dark Knight? [Assistant]: The sequel to Batman Begins is The Dark Knight. [Human]: Okay. Who is the director of The Dark Knight and any other movies from him not related to Batman?} \\
            \rowcolor{tGreen}{\fontfamily{ppl}\selectfont \textbf{\#Response\#:} Christopher Nolan was the director. He also directed insomnia and inception.} \\
            \rowcolor{tGreen}{\fontfamily{ppl}\selectfont \textbf{\#Your Judgement\#:} No} \\
            \rowcolor{tGreen}{\fontfamily{ppl}\selectfont \textbf{\#Dialogue History\#:} [Human]: Could you recommend movies similar to The Dark Knight? [Assistant]: The sequel to Batman Begins is The Dark Knight. [Human]: Okay. Who is the director of The Dark Knight and any other movies from him not related to Batman?} \\
            \rowcolor{tGreen}{\fontfamily{ppl}\selectfont \textbf{\#Response\#:} Steven Spielberg was the director. He also directed insomnia and inception.} \\
            \rowcolor{tGreen}{\fontfamily{ppl}\selectfont \textbf{\#Your Judgement\#:} Yes} \\
            \\
            \rowcolor{tPink}{\fontfamily{ppl}\selectfont You are trying to determine if the true entity in the response is replaced with a dissimilar entity.} \\
            \specialrule{0em}{1pt}{1pt}
            \rowcolor{tGreen}{\fontfamily{ppl}\selectfont <Demonstrations>} \\
            \\
            \rowcolor{tPink}{\fontfamily{ppl}\selectfont You are trying to determine if the true entity in the response is replaced with a dissimilar entity in a different entity type.} \\
            \specialrule{0em}{1pt}{1pt}
            \rowcolor{tGreen}{\fontfamily{ppl}\selectfont <Demonstrations>} \\
            \\
            {\fontfamily{ppl}\selectfont You should try your best to determine if the response contains non-factual or hallucinated information according to the above hallucination types. The answer you give MUST be ``Yes'' or ``No''.} \\
            \\
            {\fontfamily{ppl}\selectfont \textbf{\#Dialogue History\#:} <Here is the dialogue history>} \\
            {\fontfamily{ppl}\selectfont \textbf{\#Response\#:} <Here is the hallucinated response or right response>} \\
            {\fontfamily{ppl}\selectfont \textbf{\#Your Judgement\#:}} \\
            \bottomrule
	\end{tabular}
	\caption{Instruction of hallucination recognition for knowledge-grounded dialogue.}
        \vspace{-0.2cm}
	\label{tab:dialogue-eva-instruction}
\end{table*}

\begin{table*}[t]
	\small
	\centering
	\begin{tabular}{p{0.97\textwidth}}
		\toprule
            \rowcolor{LightSkyBlue1}{\fontfamily{ppl}\selectfont I want you act as a summary judge. Given a document and a summary, your objective is to determine if the provided summary contains non-factual or hallucinated information. You SHOULD give your judgement based on the following hallucination types and the world knowledge.} \\
            \\
            \rowcolor{tPink}{\fontfamily{ppl}\selectfont You are trying to determine if the summary is factual but some information cannot be directly inferred or entailed from the document.} \\
            \specialrule{0em}{1pt}{1pt}
            \rowcolor{tGreen}{\fontfamily{ppl}\selectfont \textbf{\#Document\#:} The panther chameleon was found on Monday by a dog walker in the wooded area at Marl Park. It had to be put down after X-rays showed all of its legs were broken and it had a deformed spine. RSPCA Cymru said it was an "extremely sad example of an abandoned and neglected exotic pet". Inspector Selina Chan said: "It is a possibility that the owners took on this animal but were unable to provide the care he needs and decided to release him to the wild. "We are urging potential owners of exotic animals to thoroughly research what is required in the care of the particular species before taking one on. "Potential owners need to make sure they can give their animal the environment it needs and they have the facilities, time, financial means and long-term commitment to maintain a good standard of care, as required under the Animal Welfare Act 2006." She added it was illegal to release non-native species into the wild.} \\
            \rowcolor{tGreen}{\fontfamily{ppl}\selectfont \textbf{\#Summary\#:} A chameleon that was found in a Cardiff park has been put down after being abandoned and neglected by its owners.} \\
            \rowcolor{tGreen}{\fontfamily{ppl}\selectfont \textbf{\#Your Judgement\#:} Yes} \\
            \\
            \rowcolor{tPink}{\fontfamily{ppl}\selectfont You are trying to determine if there exists some non-factual and incorrect information in the summary.  } \\
            \specialrule{0em}{1pt}{1pt}
            \rowcolor{tGreen}{\fontfamily{ppl}\selectfont <Demonstrations>} \\
            \\
            \rowcolor{tPink}{\fontfamily{ppl}\selectfont You are trying to determine if there is a factual contradiction between the summary and the document.} \\
            \specialrule{0em}{1pt}{1pt}
            \rowcolor{tGreen}{\fontfamily{ppl}\selectfont <Demonstrations>} \\
            \\
            {\fontfamily{ppl}\selectfont You should try your best to determine if the summary contains non-factual or hallucinated information according to the above hallucination types. The answer you give MUST be ``Yes'' or ``No''.} \\
            \\
            {\fontfamily{ppl}\selectfont \textbf{\#Document\#:} <Here is the test document>} \\
            {\fontfamily{ppl}\selectfont \textbf{\#Summary\#:} <Here is the hallucinated summary or right summary>} \\
            {\fontfamily{ppl}\selectfont \textbf{\#Your Judgement\#:}} \\
            \bottomrule
	\end{tabular}
	\caption{Instruction of hallucination recognition for text summarization.}
        \vspace{-0.2cm}
	\label{tab:summary-eva-instruction}
\end{table*}

\section{Details of HaluEval} \label{app-details}

The number of generated hallucinated samples for each hallucination pattern are shown in Table~\ref{tab:overall-hallucination-pattern}.

\begin{table}[t]
	\small
	\centering
	\setlength\tabcolsep{4pt}
	\begin{tabular}{l c c c c c}
		\toprule
		  \textbf{Tasks} & \textbf{\#Sample} & \textbf{P-I} & \textbf{P-II} & \textbf{P-III} & \textbf{P-IV} \\
		\midrule
		\textbf{QA} & 10000  & 2280  & 1378 & 5102 & 1240 \\
            \textbf{Dialogue}  & 10000  & 8330  & 1196 & 474 & - \\
            \textbf{Summa.}  & 10000  &  2614 & 3562 & 3824 & - \\
		\bottomrule
	\end{tabular}
	\caption{Number of generated samples for each hallucination pattern (P-I/II/III/IV). ```Summa.'' is short for summarization. ``-'' is due to that we consider three patterns in dialogue and summarization.}
        \vspace{-0.3cm}
	\label{tab:overall-hallucination-pattern}
\end{table}

\end{document}